\documentclass{article} 
\usepackage{graphicx}

\PassOptionsToPackage{numbers, compress}{natbib}

\usepackage[preprint]{neurips_2025}

\usepackage[utf8]{inputenc} 
\usepackage[T1]{fontenc}    
\usepackage{hyperref}
\usepackage{xurl}            
\usepackage{booktabs}       
\usepackage{amsfonts}       
\usepackage{nicefrac}       
\usepackage{microtype}      
\usepackage{xcolor}         
\usepackage{placeins}   
\usepackage{cleveref}
\usepackage{makecell}
\usepackage{multirow}
\usepackage{subcaption,graphicx,adjustbox}
\usepackage{pgffor}
\usepackage{enumitem} 
\usepackage{wrapfig}
\newcommand{\surveyimg}[2]{%
  \includegraphics[height=0.08\textheight,keepaspectratio]{images/survey/#1/#2}}

\newcommand{\ximg}[2]{%
  \includegraphics[height=0.10\textheight,keepaspectratio]{images/X-images/#1/#2}}

\usepackage{xspace}
\newcommand{\method}{\textsc{OpenFake}\xspace}
\usepackage{xcolor,colortbl,pifont}

\newcommand{\RICH}{\cellcolor{green!15}{\textbf{Rich}}}
\newcommand{\MOD}{\cellcolor{yellow!15}{Moderate}}
\newcommand{\NAR}{\cellcolor{red!15}{Narrow}}

\newcommand{\ACCeasy}{\cellcolor{green!15}{}}
\newcommand{\ACChard}{\cellcolor{yellow!15}{}}
\newcommand{\ACCnone}{\cellcolor{red!15}{}}

\newcommand{\starfull}{$\bigstar$}
\newcommand{\vqual}[1]{\begingroup\count0=#1\relax
  \loop\ifnum\count0>0 \starfull\advance\count0 by -1\relax\repeat\endgroup}

\usepackage{pifont}
\newcommand{\cmark}{\textcolor{green!60!black}{\ding{51}}} 
\newcommand{\xmark}{\textcolor{red!70!black}{\ding{55}}}   

\newcommand{\Low}{\cellcolor{red!25}{Low}}
\newcommand{\Medium}{\cellcolor{orange!25}{Medium}}
\newcommand{\Good}{\cellcolor{yellow!25}{Good}}
\newcommand{\High}{\cellcolor{green!25}{High}}
 
\title{\textsc{OpenFake}: An Open Dataset and Platform Toward Real-World Deepfake Detection}

\author{%
  Victor Livernoche$^{1,2}$ \space Akshatha Arodi$^{2}$ \space Andreea Musulan$^{2,3,4}$ \space Zachary Yang$^{1,2}$ \\ \textbf{Adam Salvail$^{2}$ \space Gaétan Marceau Caron$^{2}$ \space Jean-François Godbout$^{2,3}$ \space Reihaneh Rabbany$^{1,2}$} \\
  $^1$ McGill University \space
  $^2$ Mila - Quebec Artificial Intelligence Institute\\
  $^3$ Université de Montréal \space
  $^4$ IVADO \\
}

\begin{document}

\maketitle

\begin{abstract}
  Deepfakes, synthetic media created using advanced AI techniques, pose a growing threat to information integrity, particularly in politically sensitive contexts. This challenge is amplified by the increasing realism of modern generative models, which our human perception study confirms are often indistinguishable from real images.  
  Yet, existing deepfake detection benchmarks rely on outdated generators or narrowly scoped datasets (e.g., single-face imagery), limiting their utility for real-world detection.
  To address these gaps, we present \method, a large politically grounded dataset specifically crafted for benchmarking against modern generative models with high realism, and designed to remain extensible through an innovative crowdsourced adversarial platform that continually integrates new hard examples. \method comprises nearly four million total images: three million real images paired with descriptive captions and almost one million synthetic counterparts from state-of-the-art proprietary and open-source models.
  Detectors trained on \method achieve near-perfect in-distribution performance, strong generalization to unseen generators, and high accuracy on a curated in-the-wild social media test set, significantly outperforming models trained on existing datasets. Overall, we demonstrate that with high-quality and continually updated benchmarks, automatic deepfake detection is both feasible and effective in real-world settings.

\end{abstract}

\section{Introduction}

Deepfakes, realistic synthetic media generated by AI, have emerged as a serious threat to the information ecosystem \citep{CSIS_Deepfake_2023, bengio2025international}. By enabling anyone to fabricate audio-visual content of real people, deepfakes can spread false information at an unprecedented scale, eroding trust across various platforms, from social media and online content to traditional media outlets. High-profile cases (e.g., forged speeches or imagery of public figures) and the prevalence of non-consensual intimate imagery underscore the potential for harm to political stability, reputation, and public safety \cite{marchal2024generative}. Scholars have warned of an “infopocalypse” where constant exposure to fake media breeds cynicism or paranoia \cite{schick2020deepfakes}. Detecting deepfakes reliably is therefore critical to mitigate the spread of misinformation and disinformation\footnote{We adopt the term \textit{misinformation} throughout this paper to refer broadly to harmful or misleading content. Technically, \textit{misinformation} denotes false information shared without intent to deceive, while \textit{disinformation} refers to deliberately deceptive content. Our usage includes both, given the difficulty of inferring intent.}, and to restore trust in digital media. The rapid advancement of AI-generated image technologies has reached a point where distinguishing between real and synthetic images has become increasingly challenging for humans. Studies have shown that humans underperform in identifying AI-generated images, highlighting the sophistication of these generative models \citep{DIEL2024100538}. 

\begin{figure}[t]\vspace{-20pt}
    \centering  \includegraphics[width=0.98\linewidth]{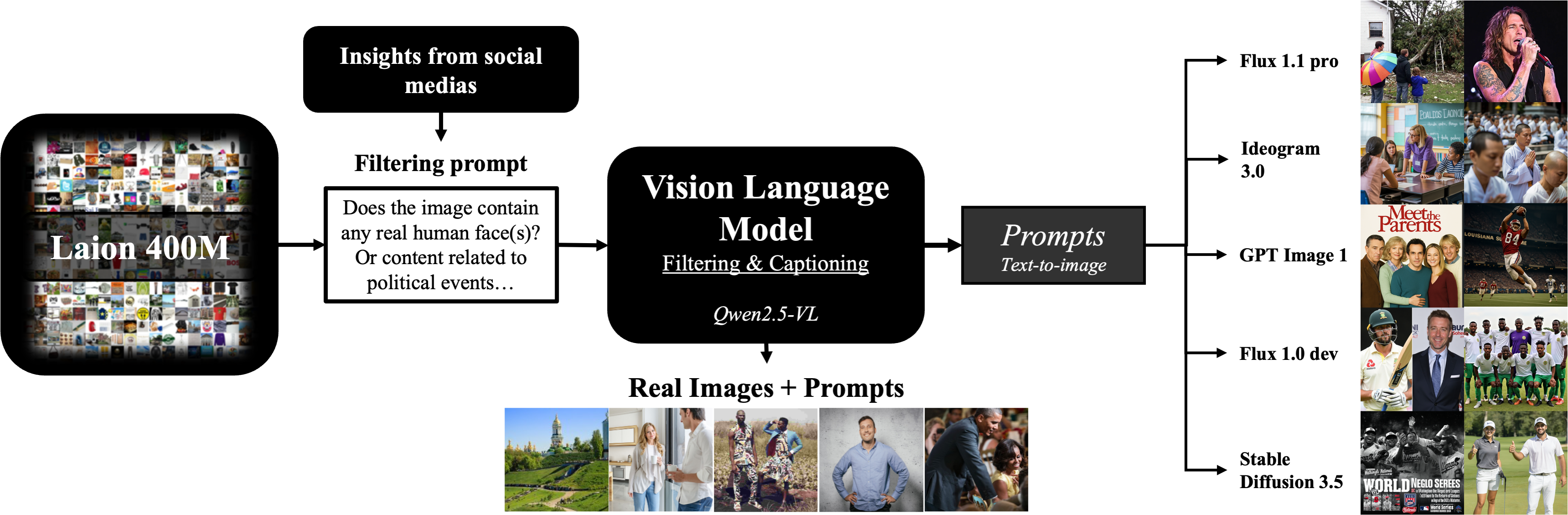}
    \caption{We begin by scraping politically relevant images from social media (e.g., X, Reddit, Bluesky), filtered by election-related hashtags. Manual investigation of these social media images helps us to design a prompt for filtering politically relevant images. A vision-language model (e.g., Qwen2.5-VL) extracts thematic captions or prompts from real images from LAION. These prompts serve dual purposes: (1) forming a large bank of real image–prompt pairs, and (2) seeding generation across a range of synthetic image models (e.g., SDv3.5, Flux, Ideogram, GPT Image 1).}
    \vspace{-1em}
    \label{fig:pipeline}
\end{figure}

The political sphere is particularly vulnerable to the risks posed by deepfakes, which can be weaponized to manipulate public opinion and undermine democratic processes \citep{bengio2025international, bengio2025singaporeconsensusglobalai, MITTechReview_Deepfakes2019}. Synthetic media have already been exploited for scams, blackmail, and targeted reputation sabotage, while the fabrication of fake historical artifacts, manipulated medical images, and staged events introduces new avenues for the spread of misinformation and societal harm \citep{Ferrara_2024, bengio2025international}. By flooding social and traditional media with convincing falsehoods, deepfakes erode public trust in news and create confusion about what is real, particularly during sensitive periods like elections \citep{IVADO_CEIMIA_AI_Democracy_2025}. Such disruptions threaten not only individual reputations and public safety, but also the legitimacy of democratic institutions and processes, with scholars warning that advances in generative AI could empower malicious actors to influence political outcomes and destabilize societies \citep{bengio2025superintelligentagentsposecatastrophic, HAMELEERS2024108096, doi:10.1177/2056305120903408}.

Based on this political motivation, we built our dataset by starting with a large-scale collection of politically relevant social media posts, including tweets, images, and videos scraped during the 2025 Canadian federal election. While this initial focus helped identify image types likely to appear in high-impact or misleading contexts, the final dataset spans a broad range of content beyond  political imagery, including everyday scenes, people, and objects. These posts serve as a foundation for analyzing real-world misinformation risks and curating image content with societal relevance. However, relying directly on social media data introduces challenges such as label contamination and unreliable ground truth, which we sought to avoid in contrast to prior work \citep{chen2023twigma}. To assess the deception risk of current generative models, we also conducted a user study measuring people’s ability to distinguish real from synthetic political images, revealing that outputs from some proprietary models may approach chance-level accuracy.

Despite significant progress in deepfake detection research \citep{li2024freqblender, shao_deepfake-adapter_2025, cozzolino2024raising}, current datasets suffer from major limitations that restrict their effectiveness in modern, real-world scenarios \citep{pal2024semitruths, chen2023twigma}. As shown in \Cref{tab:deepfake_datasets}, most established benchmarks rely on outdated generation methods; GAN-based face-swapping tools such as DeepFaceLab \citep{deepfacelab} and Face2Face \citep{thies2016face2face}. These datasets, while valuable for early detection efforts, are increasingly unrepresentative of the latest synthetic media, particularly high-fidelity diffusion and transformer-based models. Moreover, they overwhelmingly focus on single-face portraits, providing little to no real-world grounded images and neglecting the broader spectrum of image-based misinformation that floods political and social media discourse: crowd scenes, protests, disaster images, manipulated signage, or synthetic screenshots.

To address these gaps, we introduce \textsc{OpenFake}\footnote{\url{https://huggingface.co/datasets/ComplexDataLab/OpenFake}}
, a politically grounded dataset for general deepfake detection. \textsc{OpenFake} pairs large-scale real image corpora with state-of-the-art synthetic images and is designed to remain extensible through \textsc{OpenFake Arena}, a crowdsourced adversarial platform that continually contributes hard, validated examples via a CLIP-based prompt-consistency gate and scoring against a live detector. This yields a self-improving benchmark that tracks the evolution of modern generators.

By training a modest {SwinV2-Small} \citep{liu2021swin} detector on \method we produce near perfect in-distribution results on the held-out test sets (unseen images from the seen generators and their variants), as well as strong performance on images from unseen generators (see \Cref{tab:baseline_results}), with overall F1 score of 0.99 compared to the 0.88 from our strongest baseline (same model trained on \textsc{GenImage}).  
More interestingly, the model trained on \method  achieves a strong performance on a curated in-the-wild social-media test set, with a F1 score of 0.86  compared to 0.08 (\textsc{GenImage}), and 0.26 (\textsc{Semi-Truths}) 
(see \Cref{tab:wild_results}). 
This finding stands in  contrast to the prevailing pessimism around automatic deepfake detection, which is deemed as both futile and intractable, largely due to the increasing realism of generative models \citep{noauthor_artificial_2022, noauthor_fighting_nodate, noauthor_risks_nodate, noauthor_mirage_nodate, noauthor_ceo_2019, jsan14010017}. Indeed, some experts argue that distinguishing real from synthetic images will soon become impossible, which has shifted attention to watermarking \citep{wen2023treerings, watermark_min2024, liu2025image, saberi2024robustness}. However, watermarking depends on developer cooperation, consistent deployment across proprietary systems, and robustness to post-processing, and therefore cannot replace open-world detection. \method paves the way for automatic detection, demonstrating that highly realistic deepfake,   which evade detection by human eyes, can be detected with high accuracy.
In summary, our contributions are:
\begin{itemize}[leftmargin=*]
  \item   Providing \method, an \textbf{\textit{easily accessible }}dataset, which has 
    \begin{itemize}
 \item \textit{\textbf{Rich Scope}}: A large, politically relevant dataset of 3 million real images paired with extracted prompts, curated for misinformation risk and designed by studying real-world social media. 
     \item \textit{\textbf{High Realism}}: A diverse, high-quality synthetic image set spanning 963k images, generated from state-of-the-art open-source and proprietary models.
\item \textit{\textbf{Extendable}}: A scalable crowdsourcing framework (\textsc{OpenFake Arena}) for adversarial image generation, enabling continual community-driven benchmarking.

 \end{itemize}

     \item An experimental study which shows  \textbf{\textit{the weak performance of detectors trained on the currently available datasets}} on detecting realistic deepfakes, with F1 ranging between 0.0 to 0.88.  Our model trained on \method significantly outperform these baselines with a F1 of 0.99. 

    \item A human perception study showing that \textbf{\textit{images from modern proprietary generators can be imperceptible to humans,}} with accuracy in some cases dropping to near random chance (e.g., with Imagen 3). Our model trained on \method, however, achieves a near perfect performance. 
    \item A real-world feasibility study which report \textbf{\textit{a strong performance of \method on detecting real and fake images actually circulated on social media}}, based   on a carefully curated in-the-wild test set.  \method achieves a F1 score of 0.86, significantly higher that when relying on two strong contenders: 0.08 (\textsc{GenImage}), and 0.26 (\textsc{Semi-Truths}).
    
\end{itemize}


Together, \textsc{OpenFake} and \textsc{OpenFake Arena} form a robust and adaptive foundation for studying deepfakes in politically sensitive contexts, providing researchers and practitioners with the publicly available tools necessary to characterize emerging synthetic threats.

\section{Related work}

\paragraph{Synthetic image datasets.}
Despite the proliferation of generative models, existing deepfake datasets remain limited in realism, diversity, and accessibility. Early benchmarks such as FaceForensics++ \citep{faceforensicspp}, Celeb-DF \citep{li2020celeb}, and DFDC \citep{arxiv.2006_07397} rely on GAN-based face-swapping techniques and focus almost exclusively on single-person portrait videos. Even newer datasets such as ForgeryNet \citep{arxiv.2103_05630}, OpenForensics \citep{le2021openforensics}, and FFIW \citep{zhou2021face} continue to emphasize face-centric detection, with limited variation in image content or generation method \citep{cheng2024diffusion, arxiv.2403_18471, yan2024df40}. More recent image datasets have started to incorporate diffusion-based generators (e.g., Stable Diffusion, DALL·E 2, Midjourney), as seen in Fake2M \citep{lu2023seeing}, DiffusionForensics \citep{wang2023dire}, and GenImage \citep{zhu2023genimage}. However, these datasets still fall short in several ways. First, most rely on open-source models like SDv1.5 or SDv2.1 \citep{Rombach_2022_CVPR}, which, while important, do not match the visual fidelity of cutting-edge proprietary models such as Imagen 3 \citep{baldridge2024imagen} or GPT Image 1 \citep{openai2025gptimage1}. As a result, they fail to represent the modern threat landscape posed by the most deceptive fakes. Second, many datasets lack real-world grounding. Image prompts are frequently abstract, artistic, or class-based (e.g., GenImage uses classes from ImageNet-1k \citep{5206848}), failing to capture the multimodal misinformation strategies actually deployed online. Third, these datasets are static and infrequently updated, meaning they quickly become outdated as generation tools evolve. Fourth, prompts used for image generation are often withheld, making it difficult for others to reproduce, regenerate, or expand these datasets with future models. In contrast, we release a large bank of extracted prompts along with the images, which enables researchers to extend the dataset. Finally, accessibility remains a persistent issue. Many datasets require downloading large zip archives via Google Drive or web links, making them difficult to integrate into new pipelines. In contrast, \textsc{OpenFake} is fully hosted on the HuggingFace Hub in streaming-friendly Parquet format, enabling scalable access and evaluation, which should help the community develop new detection tools. \Cref{tab:deepfake_datasets} highlights these differences in model coverage, dataset scope, prompt extensibility, and access modality.

\begin{table}[h]
\centering
\resizebox{\textwidth}{!}{
\begin{tabular}{lccccccccc}
{Dataset} & {Year} & {Fakes} & {Reals} & {Extends} & {Content Scope} & {Realism} & {Access} & {Methods} & {Most recent model}\\
\midrule
FaceForensics++ \citep{faceforensicspp}        & 2019 & 5K        & 1K        & \xmark & \NAR & \Low    & \ACChard\;Hard & 4   & Face2Face (2016) \\
Celeb-DF \citep{li2020celeb}                   & 2020 & 5K+       & 590       & \xmark & \NAR & \Low    & \ACChard\;Hard & 1   & DeepFaceLab (2020) \\
DFDC \citep{arxiv.2006_07397}                  & 2020 & 100K+     & 20K+      & \xmark & \NAR & \Low    & \ACChard\;Hard & 8   & DeepFaceLab (2020) \\ 
ForgeryNet \citep{arxiv.2103_05630}            & 2021 & 1.5M      & 1.5M      & \xmark & \NAR & \Low    & \ACChard\;Hard & 15  & DeepFaceLab (2020) \\ 
FFIW \citep{zhou2021face}                      & 2021 & 10K       & 10K       & \xmark & \NAR & \Low    & \ACChard\;Hard & 3   & DeepFaceLab (2020) \\ 
OpenForensics \citep{le2021openforensics}      & 2021 & 100K      & 100K      & \xmark & \NAR & \Low    & \ACChard\;Hard & 3   & GAN (2020) \\ 
DeepFakeFace \citep{arxiv.2309_02218}          & 2023 & 90K       & 30K       & \xmark & \NAR & \Medium & \ACChard\;Hard & 3   & SD v1.5 (2022) \\
Fake2M \citep{lu2023seeing}                    & 2023 & \textbf{2M}        & 0         & \xmark & \MOD & \Medium & \ACCeasy\;\textbf{Easy} & 3   & SD v1.5 (2022) \\
DiffusionForensics \citep{wang2023dire}        & 2023 & 570K      & 140K      & \xmark & \MOD & \Medium & \ACChard\;Hard & 8   & iDDPM (2021) \\
DMDetection \citep{corvi2023detection}         & 2023 & 200K      & 200K      & \xmark & \MOD & \Medium & \ACChard\;Hard & 3   & DALL·E 2 (2022) \\
GenImage \citep{zhu2023genimage}               & 2023 & 1.3M      & 1.33M     & \xmark & \MOD & \Good   & \ACChard\;Hard & 5   & Midjourney 5 (2023) \\
TWIGMA \citep{chen2023twigma}                  & 2023 & 800K      & 0         & \xmark & \RICH& \Medium & \ACCnone\;Unavailable & -- & -- \\
DiffusionDeepfake \citep{arxiv.2404_01579}     & 2024 & 100K      & 94K       & \xmark & \NAR & \Good   & \ACChard\;Hard & 2   & Midjourney (2024) \\ 
DF40 \citep{yan2024df40}                       & 2024 & 1M+       & 1.5K      & \xmark & \NAR & \Medium & \ACChard\;Hard & \textbf{40} & PixArt-$\alpha$ (2024) \\ 
DiffusionFace \citep{arxiv.2403_18471}         & 2024 & 600K      & 30K       & \xmark & \NAR & \Good   & \ACChard\;Hard & 11  & SD v2.1 (2022) \\ 
DiFF \citep{cheng2024diffusion}                & 2024 & 500K      & 23K       & \xmark & \NAR & \Good   & \ACChard\;Hard & 13  & Midjourney 5 (2023) \\
Semi-Truths \citep{pal2024semitruths}          & 2024 & 1.34M     & 26K       & \xmark & \MOD & \Good   & \ACCeasy\;\textbf{Easy} & 8   & Stable Diffusion XL (2023) \\
\midrule
\textsc{OpenFake} (Ours)                       & 2025 & 963K & \textbf{3M} & \cmark & \RICH & \textbf{\High}  & \ACCeasy\;\textbf{Easy} & 18  & Imagen 4.0 \textbf{(2025)} \\
\bottomrule
\end{tabular}
}
\vspace{0.7em}
\caption{\textbf{Compared to current public deepfake datasets, OpenFake uniquely combines rich scope, high realism, large real sample count, easy access, and extensibility.} “Fakes” and “Reals” count individual media items (images or videos; units omitted for brevity). 
\textbf{Content Scope}: Narrow (face-focused); Moderate (diverse but limited); Rich (broad, internet-like variety). 
\textbf{Access}: \ACCnone\ Unavailable; \ACChard\ Hard (public but cumbersome); \ACCeasy\ Easy (clean, ready-to-use hosting). 
\textbf{Realism}: qualitative fidelity of synthetic content — \Low, \Medium, \Good, \High. 
\textbf{Extendable}: availability of prompts/metadata enabling seamless dataset expansion. 
}
\vspace{-1em}
\label{tab:deepfake_datasets}
\end{table}

\paragraph{Deepfake detection methods.} 
Early detection approaches relied on convolutional neural networks trained on known forgery artifacts, such as blending boundaries or low-level inconsistencies in the images \citep{afchar2018mesonet, faceforensicspp, 10363477}. While effective in-domain, these models struggle to generalize across generation techniques \citep{ojha2023fakedetect}. As diffusion and transformer-based models reduce such artifacts, recent work has focused on semantic-level signals and frequency-domain cues \citep{durall2020watch, liu2021spatial, frank2020leveraging, qian2020thinking}. CLIP-based detection \citep{cozzolino2024raising, clippingdeception} has emerged as a promising direction, leveraging large-scale vision-language pretraining to improve robustness. Other advances include domain-adaptive feature learning \citep{shao_deepfake-adapter_2025, jia2024can}, zero-shot detectors \citep{lin2024robust}, and hybrid approaches that blend local artifact patterns with global semantic reasoning \citep{li2024freqblender, zhou2024capture, ma2025specificity}. Despite progress, the rapid pace of generative model development continues to outstrip detection capabilities, motivating adaptive benchmarks like \textsc{OpenFake Arena} to assess robustness in a dynamic, adversarial setting. 

\section{Case Study: Real-World Misinformation and Human Limits}

\label{sec:social_media}

\begin{figure}[h]
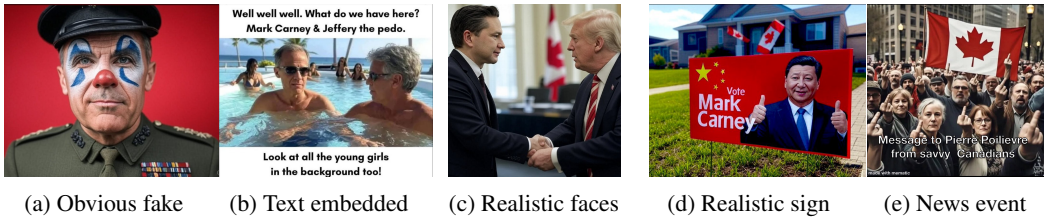

\centering
\begin{subfigure}[b]{0.19\textwidth}
\centering
\ximg{deepfakes}{11.png}
\caption{Obvious fake}
\end{subfigure}
\hfill
\begin{subfigure}[b]{0.19\textwidth}
\centering
\ximg{deepfakes}{17.png}
\caption{Text embedded}
\end{subfigure}
\hfill
\begin{subfigure}[b]{0.19\textwidth}
\centering
\ximg{deepfakes}{35.png}
\caption{Realistic faces}
\end{subfigure}
\hfill
\begin{subfigure}[b]{0.19\textwidth}
\centering
\ximg{deepfakes}{14.png}
\caption{Realistic sign}
\end{subfigure}
\begin{subfigure}[b]{0.19\textwidth}
\centering
\ximg{deepfakes}{8.png}
\caption{News event}
\end{subfigure}
\caption{Examples of deepfake images collected from X depicting various types of fabricated scenarios involving Canadian political figures and events.}
\label{fig:x_deepfake_examples}
\end{figure}

Social media platforms have become critical channels for political discourse, and consequently, for amplifying deepfake disinformation. This raises a key question: how are deepfakes actually deployed within political conversations? As part of a subsequent study, and to investigate how deepfakes are used in political conversations and to later evaluate our detector in a realistic setting, we collected images from X, Reddit, and Bluesky. The collection spans the period immediately before and during the 2025 Canadian federal election, enabling analysis of synthetic media in a high-stakes, time-sensitive information environment.


We manually examined over 2000 randomly sampled images collected during a 72-hour period to better understand the types of visuals circulating on social media, identifying 163 deepfakes (see \Cref{sec:wild} for more details). Many of these prominently featured political figures. Fabricated scenarios involving leading candidates distorted public perception, and even when some deepfakes were clearly artificial (\Cref{fig:x_deepfake_examples}a), they reinforced existing biases more effectively than textual misinformation alone \citep{ecker2022psychological, Hameleers03032020, doi:10.1177/2056305120903408}. When photorealistic deepfakes aligned with viewers’ prior beliefs, the risk was higher, as illustrated in \Cref{fig:x_deepfake_examples}c. Beyond portraits, misinformation extended to political symbols, banners, and manipulated depictions of protests or disasters (\Cref{fig:x_deepfake_examples}d,e), often paired with misleading text (\Cref{fig:x_deepfake_examples}b). These findings inform the construction of our dataset: to support generalizable detection, a deepfake benchmark must move beyond faces to capture the full breadth of misleading visual content. While this dataset serves here to illustrate the variety of real-world deepfakes, in \Cref{sec:wild} we also use it as a small in-the-wild evaluation set. Importantly, none of these images or their captions are included in training or generation, ensuring a clean separation between evaluation and benchmark design. 


\subsection{Human Perception Study}
\label{sec:survey}

\begin{figure}[t]
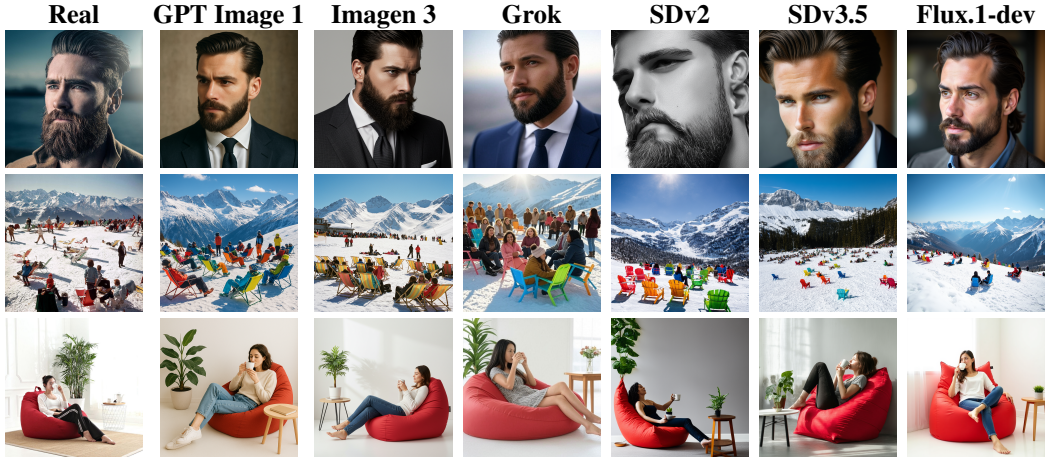

\centering
\setlength{\tabcolsep}{2pt}      
\renewcommand{\arraystretch}{0.8}

\begin{tabular}{ccccccc}
\textbf{Real} & \textbf{GPT Image 1}  & \textbf{Imagen 3} & \textbf{Grok} & \textbf{SDv2} & \textbf{SDv3.5} & \textbf{Flux.1-dev}\\
\surveyimg{real}{image_13.jpg}    & 
\surveyimg{4o}{image_13.png}    & \surveyimg{Imagen}{image_13.jpg} & \surveyimg{Grok}{image_13.png} & \surveyimg{SD2}{image_13.png} & \surveyimg{SDv3.5}{image_13.png} & \surveyimg{Flux.1-dev}{image_13.png}\\
\surveyimg{real}{image_21.jpg}    & 
\surveyimg{4o}{image_21.png}    & \surveyimg{Imagen}{image_21.jpg} & \surveyimg{Grok}{image_21.png} & \surveyimg{SD2}{image_21.png} & \surveyimg{SDv3.5}{image_21.png} & \surveyimg{Flux.1-dev}{image_21.png}\\
\surveyimg{real}{image_22.jpg}     & \surveyimg{4o}{image_22.png}     & \surveyimg{Imagen}{image_22.jpg}  & \surveyimg{Grok}{image_22.png}  & \surveyimg{SD2}{image_22.png}  & \surveyimg{SDv3.5}{image_22.png}  & \surveyimg{Flux.1-dev}{image_22.png}
\end{tabular}
\caption{Examples of deepfakes from each model used in the survey with their respective real image.}
\vspace{-1em}
\label{fig:survey_examples}
\end{figure}

To assess the difficulty of detecting deepfakes generated by different models, we conducted a simple human study. A total of 100 participants completed the survey, each viewing a randomized set of 24 images. The set consisted of 12 real photographs and 12 synthetic images—2 from each of the six generation models: GPT Image 1 (OpenAI) \citep{openai2025gptimage1}, Imagen 3 (Google) \citep{baldridge2024imagen}, Grok 2 (xAI), Flux.1.0-dev (Black Forest Labs) \citep{flux2024}, Stable Diffusion 2.1 \cite{Rombach_2022_CVPR}, and Stable Diffusion 3.5. All synthetic images were generated from the same prompts as their real counterparts, using the text automatically extracted by our pipeline described in \Cref{sec:data_collection}. 
Each prompt was only shown once to a given participant. This ensured that responses reflected a fair and diverse exposure across the dataset. In total, the survey contained 168 unique images. The survey description, code, and images are available at \url{https://github.com/vicliv/deepfake-survey}.



\begin{table}[h]
\centering
\resizebox{\linewidth}{!}{%
\begin{tabular}{lccccccccc}
\makecell{{Source}\\} & \makecell{{Release}\\} & \makecell{{Access}\\ } & \makecell{{\textbf{Humans}}\\ } & \makecell{{CLIP-D-10k+}\\ } & \makecell{{Corvi2023}\\} & \makecell{{Fusion}\\{(CLIP+Corvi)}} &
\makecell{{SwinV2}\\{\small(GenImage)}} &
\makecell{{SwinV2}\\{\small(Semi-Truths)}} &
\makecell{{SwinV2}\\\textbf{(\method)}} \\
\midrule
\rowcolor{green!20}
Real & --- & --- & 0.718 & 0.479 & 0.000 & 0.062 & 1.000 & 0.500 & \textbf{1.000 }\\
\midrule
Imagen 3 & 2024 & Proprietary & 0.490 & 0.458 & 0.708 & 0.667 & 0.625 & 1.000 & 1.000 \\
GPT Image 1 & 2025 & Proprietary & 0.684 & 0.458 & 0.500 & 0.458 & 0.750 & 1.000 & 1.000 \\
Flux.1.0-dev & 2024 & Open & 0.689 & 0.562 & 0.792 & 0.812 & 0.917 & 1.000 & 1.000 \\
SDv3.5 & 2024 & Open & 0.709 & 0.521 & 1.000 & 0.938 & 0.750 & 1.000 & 1.000 \\

SDv2.1 & 2022 & Open & 0.879 & 0.646 & 1.000 & 0.938 & 0.917 & 1.000 & 1.000 \\
Grok 2 & 2024 & Proprietary & 0.811 & 0.771 & 1.000 & 0.938 & 0.208 & 1.000 & 1.000 \\
\midrule
\midrule
\multicolumn{3}{l}{Overall Accuracy} & 0.714 & 0.524 & 0.417 & 0.427 & 0.804 & 0.750 & \textbf{1.000} \\
\bottomrule
\end{tabular}
}
\vspace{0.7em}
\caption{Survey results showing human true positive rate (TPR) for synthetic images and true negative rate (TNR) for real images. A score near 0.50 indicates chance-level performance.  SwinV2 trained on \textsc{OpenFake} achieves high accuracy on its in-distribution generators.} 
\vspace{-1em}
\label{tab:survey_results}
\end{table}


The results in \Cref{tab:survey_results} highlight key insights into synthetic image realism and human perception. Imagen 3 from Google achieved the lowest human accuracy (48.5\%), equivalent to random guessing. GPT Image 1, OpenAI’s recent model, was similarly deceptive, with nearly one-third of its images undetected. In contrast, Stable Diffusion 2.1 (SDv2.1, which is widely used in current benchmarks, had the highest detection rate (87.9\%) and was easily flagged by all detectors. These findings suggest that while open-source models remain relatively easy to spot, advanced proprietary models demonstrate exceptional realism and consistency. This underscores the need to include such models in benchmarks, as current deepfake detection datasets trained on older or open-source models fail to capture the new quality standards set by proprietary systems.

However, this perceptual difficulty is not limited to humans. Baseline deepfake detectors also failed to consistently identify these advanced fakes. The CLIP-D-10k+ method \citep{cozzolino2024raising}, which fine-tunes a linear classifier on CLIP embeddings, performed close to random on several models (e.g., 45.8\% on Imagen 3, and GPT Image 1), and failed to distinguish real images entirely (47.9\% TNR). Similarly, the Corvi2023 method \citep{corvi2023detection}, which uses curated handcrafted features for detection, fared better on some open-source models (e.g., 100\% on SDv3.5), but completely failed on real images (0.0\% TNR) and newer proprietary content like GPT Image 1 (50.0\%). In contrast, our SwinV2 baseline, trained directly on a curated mix of real and synthetic images from models included in our dataset, achieved 100\% overall accuracy and perfect performance on its in-distribution models. While this result demonstrates that deep networks can learn to detect even the most realistic fakes with sufficient supervision, it also highlights the brittleness of existing methods when faced with novel or unseen generative sources. The two additional SwinV2 baselines confirm this pattern: models trained on smaller or differently distributed data (GenImage and Semi-Truths) showed weaker generalization to modern deepfakes, with accuracy dropping to 80.4\% and 75.0\%, respectively. These results reinforce the importance of training on a large, diverse and higher-quality dataset like OpenFake to ensure robustness against emerging generation techniques.



\section{Dataset Overview \& Collection}
\label{sec:data_collection}

\textsc{OpenFake} combines a large repository of real images with a diverse collection of high-quality synthetic counterparts generated by multiple state-of-the-art models. In Appendix~\ref{ap:stats}, \Cref{tab:dataset_stats} presents in details the key statistics of the dataset. \Cref{fig:tsne} offers a qualitative view of the underlying distribution of some of the images. The substantial overlap between real and synthetic samples in the CLIP feature space highlights their semantic alignment, suggesting that synthetic images effectively mimic the distribution of real-world content. This shows that the prompt generation pipeline is working as intended. \Cref{fig:pipeline} presents an overview of the data collection and generation process.

\paragraph{Real images.}
We extract metadata from the LAION-400M dataset \citep{schuhmann2021laion}, which we selected due to its broad representation of internet-sourced images—the same domain where visual misinformation typically circulates. Additionally, this dataset was likely included in the training data of the text-to-image models used to generate the synthetic images, which should theoretically make it more difficult for detectors to distinguish between real and fake images. More importantly, these images preserve real-world compression artifacts, which are crucial for training detectors that operate in the wild. While LAION may contain some synthetic images, we expect this contamination to be minimal, as the dataset primarily consists of content from 2014–2021, before the public release of diffusion models in 2022. After scraping LAION, we filter image–caption pairs using a vision–language model (Qwen2.5‑VL \citep{bai2025qwen2}). As described by the prompt used to query the model in \Cref{ap:filtering}, an image is retained if it is identified as depicting either (i) real human faces or (ii) politically salient or newsworthy events. For every retained image, we generate a more detailed caption to use as prompt input for text-to-image models. These $3\,\mathrm{M}$ prompts are also publicly released and form the basis of the prompts shown to users of our crowdsourcing platform \textsc{OpenFake Arena}.

We filtered real images using Qwen2.5-VL, selected for its trade-off between speed and quality. To prevent detection shortcuts, we excluded LAION images smaller than 512\(\times\)512 pixels from the released \texttt{train/test} sets, as lower resolutions introduced compression artifacts that made detection artificially easier. Full prompts used for filtering and captioning are provided in \Cref{ap:filtering}, and additional details on generation and compute resources are in \Cref{ap:details}.

\paragraph{Synthetic images.}
We generated images from a diverse set of state-of-the-art generators using samples from our prompt bank: Stable Diffusion 1.5/2.1/XL/3.5 \citep{Rombach_2022_CVPR}, Flux 1.0-dev/1.1-Pro/Schnell \citep{flux2024}, Midjourney v6/v7 \citep{midjourney-v61,midjourney-v7}, DALL\textperiodcentered E~3 \citep{dalle3}, Imagen 3/4 \citep{baldridge2024imagen,imagen4-vertex}, GPT Image 1 \citep{openai2025gptimage1}, Ideogram 3.0 \citep{ideogram-3}, Grok-2 \citep{grok-image}, HiDream-I1 \citep{hidream-i1-hf,hidream-i1-git}, Recraft v3 \citep{recraft-v3-blog}, Chroma \citep{chroma-hf}, and 10 community variants (Finetuned or LoRA) of Stable Diffusion 1.5/XL and Flux-dev. All images are produced at $\sim$1\,MP resolution with varied aspect ratios (9:16, 16:9, 1:1, 2:3, 3:4, \emph{etc.}), mirroring common social-media formats. Because several proprietary sources impose “non-compete” clauses, those subsets are released under a non-commercial license.

\paragraph{Splits and accessibility.}
We construct the \texttt{train/test} split by sampling 1,000 images per generative model (with the exception of out-of-distribution models, which contribute between 200 and 600 images) along with the corresponding number of real images, yielding a test set of roughly 60,000 images (about 3\% of the dataset assuming balanced classes). The remaining images are allocated to the training set. To ensure balance, each model is equally represented in the \texttt{test} split, and real images are matched accordingly. The rest of the real images and prompts are provided in a CSV file, with the real images accessible through their URLs. As \textsc{OpenFake Arena} expands and more synthetic images are collected, additional real images will be incorporated to preserve parity between real and synthetic domains in the \texttt{train/test} splits.
All assets are hosted on the HuggingFace Hub.
All images have their associated prompts and model name as metadata, which can be used for model attribution.




\section{Crowdsourced Adversarial Platform }
\label{sec:arena}
Generative and detection models co-evolve: advances in generation demand stronger detectors, which in turn promote new generation models. To keep benchmarks relevant amid rapid progress, we introduce \textsc{OpenFake Arena}: a crowdsourced platform where users generate synthetic images to fool a live detector. Successful examples are added to the benchmark, enabling sustained evaluation.

\textsc{OpenFake Arena}\footnote{\url{https://huggingface.co/spaces/CDL-AMLRT/OpenFakeArena}} is designed as a web-based interactive game to encourage wide participation. Each round begins with a prompt sampled from our bank of over 3 million. Users respond by generating a synthetic image using any generative model or editing tool that aligns with the prompt. A CLIP-based similarity gate verifies prompt-image alignment.  If the image passes this check, it is evaluated by a detector trained on the \textsc{OpenFake} dataset. If the detector misclassifies the synthetic image as real, the user earns a point and the image is added to the benchmark.

The Arena features a real-time leaderboard to gamify the experience and incentivize participation. The detector is periodically retrained with newly collected data, enabling continual improvement. Submitted images are periodically reviewed. This human-in-the-loop setup transforms model drift from a challenge into a feature, allowing the benchmark to evolve organically alongside the state of generative models. Implementation details and screenshots of the arena are in \Cref{ap:arena}. 

\section{Baseline Detector Benchmarks}
\label{sec:baselines}
We evaluate a selection of deepfake detectors on the \textsc{OpenFake} dataset, with the goal of assessing how well existing models generalize to modern synthetic media, especially high-quality images from diffusion and transformer-based models.

\paragraph{Benchmark models.}  
Our primary detector is \textbf{SwinV2-Small} \citep{liu2021swin}, a hierarchical vision transformer that has achieved state-of-the-art results on large-scale classification tasks. We adopt it as the backbone for our supervised detector, trained on \textsc{OpenFake}. In addition, we evaluate two SwinV2 variants trained on external datasets (GenImage and Semi-Truths, chosen based on relevance from \Cref{tab:dataset_stats}), a \textbf{ConvNeXt} baseline trained on DRCT taken from \citep{chen2024drct}, and an \textbf{EfficientNet-B4} baseline trained on FaceForensics++ from \citep{Yan2023_DeepfakeBench_W4383468826}. For semi-supervised detection, we include the \textbf{CLIP-Based Synthetic Image Detector} \citep{cozzolino2024raising}, which applies a linear probe over CLIP embeddings with minimal training data. For zero-shot detection, we use \textbf{InternVL} \citep{chen2024internvl} directly without finetuning.\footnote{We use the InternVL3-38B; 
Code for all scripts and baselines \url{https://github.com/vicliv/OpenFake}
} 
Finally, we also test the handcrafted detector of \citet{corvi2023detection} and a hybrid fusion baseline that averages predictions from CLIP and Corvi2023. Together, these baselines cover a diverse range of architectures, training regimes, and prior benchmarks, allowing us to compare detectors trained on OpenFake against both legacy and contemporary approaches.

\begin{table}[h]
\centering
\small
\resizebox{\textwidth}{!}{%
\setlength{\tabcolsep}{5pt}
\begin{tabular}{l ccccccc c}

&
\makecell{\small\textbf{\method}\\{SwinV2}} &
\makecell{\small{GenImage}\\{SwinV2}} &
\makecell{\small{S.-Truths}\\{SwinV2}} &
\makecell{\small DRCT\\{ConvNeXt}} &
\makecell{\small FF++\\{EffNet-B4}} &
\makecell{\small{CLIP-}\\{D-10k+}} &
\makecell{\small{DMD}\\Corvi'23} &
\makecell{\small{InternVL-3}\\{(zero-shot)}} \\
\midrule
\midrule
\rowcolor{green!20}
Real  (TNR) &\textbf{ 0.995 }& 0.955 & 0.689 & 0.777 & 0.516 & 0.703 & 0.998 & 0.431 \\
\midrule\midrule
SD 1.5 & 1.000 & 0.936 & 1.000 & 0.447 & 0.529 & 0.579 & 0.000 & 0.849 \\
SD 2.1 & 1.000 & 0.998 & 0.999 & 0.482 & 0.453 & 0.717 & 0.011 & 0.900 \\
SD XL  & 1.000 & 0.956 & 1.000 & 0.426 & 0.507 & 0.438 & 0.001 & 0.814 \\
SD 3.5 & 1.000 & 0.982 & 1.000 & 0.324 & 0.466 & 0.406 & 0.000 & 0.796 \\
\midrule
Flux 1.0 Dev         & 1.000 & 0.967 & 0.999 & 0.290 & 0.450 & 0.401 & 0.005 & 0.748 \\
Flux-1.1-Pro         & 1.000 & 0.315 & 0.975 & 0.319 & 0.467 & 0.596 & 0.000 & 0.722 \\
Flux-1.0-Schnell     & 0.999 & 1.000 & 0.998 & 0.289 & 0.476 & 0.503 & 0.000 & 0.803 \\
\midrule
Midjourney 6         & 1.000 & 0.090 & 0.949 & 0.166 & 0.486 & 0.100 & 0.000 & 0.884 \\
Midjourney 7         & 0.994 & 0.952 & 0.997 & 0.264 & 0.484 & 0.404 & 0.001 & 0.961 \\
DALL\textperiodcentered E~3 & 0.995 & 0.238 & 0.927 & 0.461 & 0.543 & 0.394 & 0.000 & 0.983 \\
GPT Image 1          & 0.998 & 0.772 & 0.983 & 0.402 & 0.442 & 0.384 & 0.005 & 0.932 \\
Ideogram 3.0         & 1.000 & 0.993 & 1.000 & 0.254 & 0.481 & 0.414 & 0.001 & 0.844 \\
Imagen 3.0           & 0.999 & 0.962 & 0.998 & 0.237 & 0.461 & 0.286 & 0.005 & 0.784 \\
Imagen 4.0           & 0.996 & 0.948 & 0.996 & 0.228 & 0.459 & 0.359 & 0.003 & 0.796 \\
Grok 2               & 1.000 & 0.142 & 0.963 & 0.383 & 0.463 & 0.303 & 0.000 & 0.805 \\
HiDream-I1 Full      & 1.000 & 0.976 & 0.993 & 0.332 & 0.440 & 0.485 & 0.000 & 0.789 \\
Chroma               & 0.992 & 0.980 & 0.995 & 0.451 & 0.435 & 0.298 & 0.003 & 0.726 \\
\midrule
\textcolor{blue}{\textit{Ideogram 2.0}} & 0.993 & 0.997 & 1.000 & 0.234 & 0.482 & 0.777 & 0.000 & 0.865 \\
\textcolor{blue}{\textit{Lumina}}       & 1.000 & 1.000 & 1.000 & 0.494 & 0.355 & 0.720 & 0.028 & 0.983 \\
\textcolor{blue}{\textit{Frames}}       & 0.968 & 0.816 & 1.000 & 0.368 & 0.392 & 0.920 & 0.000 & 0.912 \\
\textcolor{blue}{\textit{Halfmoon}}     & 0.995 & 0.953 & 1.000 & 0.263 & 0.353 & 0.632 & 0.000 & 0.832 \\
\textcolor{blue}{\textit{Recraft v2}}   & 0.972 & 0.699 & 1.000 & 0.379 & 0.443 & 0.248 & 0.004 & 0.929 \\
\textcolor{blue}{\textit{Recraft v3}}   & 0.701 & 0.288 & 0.997 & 0.364 & 0.497 & 0.430 & 0.002 & 0.912 \\
\midrule
Average TPR           & 0.988 & 0.823 & 0.992 & 0.354 & 0.475 & 0.443 & 0.003 & 0.827 \\
\midrule\midrule
{\footnotesize Overall} F1            & \textbf{0.992} & 0.881 & 0.861 & 0.449 & 0.485 & 0.509 & 0.005 & 0.697 \\
\small{Overall} ROC AUC       &\textbf{1.000} & 0.926 & 0.960 & 0.616 & 0.493 & 0.600 & 0.487 & 0.629 \\
\small{Overall} PR AUC        & \textbf{1.000} & 0.949 & 0.952 & 0.613 & 0.493 & 0.600 & 0.488 & 0.586 \\
\bottomrule
\end{tabular}%
}
\vspace{0.7em}
\caption{Performance comparison on \textsc{OpenFake} across detectors trained on different datasets. Finetuned (FT) and LoRA variants are grouped under their respective base generators. Generators shown in \textcolor{blue}{blue} are out-of-distribution for all detectors. SwinV2 trained on \textsc{OpenFake} consistently outperforms others on unseen generators, while most alternative detectors exhibit high false positive rates (misclassification of real images).}
\vspace{-1em}
\label{tab:baseline_results}
\end{table}

\paragraph{Results on \textsc{OpenFake}.}
As shown in \Cref{tab:baseline_results}, the SwinV2 trained on \textsc{OpenFake} is near-perfect, confirming that modern classifiers are highly reliable when trained on the evaluation distribution. Separately, we evaluate robustness to compression artifacts and, with artifact-matching augmentation, the same model attains an F1 of 0.992 on a fully compressed test set (see \Cref{ap:artifacts} for more details). The SwinV2 trained on \textsc{GenImage} is the next strongest and handles many shared open-source families, but degrades on newer proprietary generators (Grok 2, Midjourney 6, Flux-1.1 pro, DALL\textperiodcentered E~3), reflecting a distribution gap in the fake images. The SwinV2 trained on \textsc{Semi-Truths} achieves high TPRs yet misclassifies many real images, which is consistent with its training data focused on edits rather than full generation. Legacy or narrow baselines (ConvNeXt/DRCT, EffNet-B4/FF++) underperform, CLIP-D-10k+ is middling, and Corvi2023 largely predicts “real,” while zero-shot InternVL is better than older baselines but still trails supervised models. Overall, robust performance requires training on the correct, broad, and up-to-date image distribution.

\paragraph{Transferability to unseen models.}
Using the out-of-distribution generators in \Cref{tab:baseline_results} (blue rows), which were collected from public web sources rather than generated by us, the SwinV2 trained on \textsc{OpenFake} shows the strongest transfer while keeping a high true-negative rate on real images. The SwinV2 trained on \textsc{GenImage} is competitive on several open-source families but lags on newer proprietary models, and the \textsc{Semi-Truths} model attains high TPRs yet mislabels many real images, so its apparent OOD gains are not reliable. Cross-benchmark tests as seen in \Cref{tab:swin_cross_bench} (Appendix~\ref{ap:cross_bench}) reinforce this: when evaluated on \textsc{GenImage}, the \textsc{OpenFake} model substantially outperforms the \textsc{Semi-Truths} model (Accuracy 0.849 vs. 0.613; F1 0.836 vs. 0.714), and when evaluated on \textsc{Semi-Truths} it exceeds the \textsc{GenImage} model (Accuracy 0.920 vs. 0.865; F1 0.947 vs. 0.907), despite being out-of-domain in both cases. This suggests that while \textbf{dataset coverage is the main driver of transferability}, there is also some degree of cross-generator generalization, likely because different models share subtle artifacts; the broader the training distribution, the more likely a detector can recognize previously unseen generators.


\subsection{Detector in the wild}
\label{sec:wild}
Performance on benchmarks often differs from performance in real-world settings. For deepfake detection, evaluation in the wild is particularly challenging. Manually labelling real images is not always straightforward, since, as discussed in \Cref{sec:survey}, humans struggle to identify high-quality fakes. Real images are easier to validate because their authenticity can often be established through provenance cues such as credits from a reputable source, multiple photos of the same event, consistent backgrounds, or camera metadata. Images without any such evidence can be discarded as uncertain. The risk, however, is that the benchmark becomes trivial, with fakes limited to the easiest cases. With an experienced labeler, the aid of reverse image search, and contextual text (which may explicitly indicate AI generation), more difficult deepfakes can be identified. Using this approach, we constructed a small evaluation set of social media images, described in \Cref{sec:social_media}, containing 1,057 real images and 163 labeled as deepfakes by us, and compared the performance of our SwinV2 baseline trained on \textsc{OpenFake}, \textsc{GenImage}, and \textsc{Semi-Truths} (\Cref{tab:wild_results}), since these were the only competitive baselines from \Cref{tab:baseline_results}.

\begin{table}[h]
\centering
\small
\begin{tabular}{lccc}
\toprule
\textbf{Metric} &
\makecell{\textbf{Train}\\\textsc{OpenFake}} &
\makecell{\textbf{Train}\\\textsc{GenImage}} &
\makecell{\textbf{Train}\\\textsc{Semi-Truths}} \\
\midrule
TNR       & 0.976 & 0.998 & \color{red}{0.220} \\
TPR       & 0.865 &\color{red} 0.043 & 0.908 \\
\midrule
Accuracy  & \textbf{0.962} & 0.871 & 0.312 \\
F1 Score  & \textbf{0.857} & 0.081 & 0.261 \\
ROC--AUC  & \textbf{0.978} & 0.557 & 0.634 \\
\bottomrule
\end{tabular}
\vspace{0.7em}
\caption{Generalization of SwinV2 detectors trained on different benchmarks when evaluated on an \emph{in-the-wild} social-media set (1,057 real, 163 fake; see \Cref{sec:social_media}). Metrics include TNR (real) and TPR (fake). Training on \textsc{OpenFake} yields balanced performance, while \textsc{GenImage} and \textsc{Semi-Truths} show strong class biases.}
\label{tab:wild_results}
\end{table}

The detector trained on \textsc{OpenFake} shows encouraging results. It produces very few false positives, meaning real images are rarely misclassified as deepfakes, while still identifying 86.5\% of the fakes. Although the evaluation set is small and not without limitations—labels were verified, but some of the most difficult cases may have been discarded during curation (out of roughly 2000 candidate images, many were removed, including irrelevant real samples such as screenshots of text or drawings); the results strongly suggest that \textsc{OpenFake} offers superior real-world applicability compared to existing datasets. This is a promising outcome for the deepfake detection community. Expanding generator coverage and incorporating image edits, rather than only fully generated images, could further improve performance and move closer to practical, reliable detection systems.

\section{Conclusion}
\label{sec:conclusion}

We introduced \textsc{OpenFake}, a politically grounded benchmark built from three million real images paired with nearly one million high-quality synthetic counterparts, and extended it with \textsc{OpenFake Arena}, a crowdsourced adversarial platform for continual updates. Our human perception study confirmed that recent proprietary generators often fool users, while detectors trained on older datasets fail against these models. In contrast, detectors trained on \textsc{OpenFake} achieved strong in-distribution performance and promising results on a curated in-the-wild set of social-media images, suggesting that reliable detection of deepfakes is attainable outside controlled benchmarks. 

While performance on some proprietary or lower-quality sources remains uneven, the path forward is clear: expanding generator coverage and broadening real image diversity (e.g., camera types, capture conditions) to further improve robustness. By combining high-fidelity benchmarking with community-driven adversarial submissions, our framework aims to narrow the gap between generation and detection, equipping researchers and practitioners with tools to confront emerging misinformation threats in real time.

\section{Reproducibility Statement}
We have made all components necessary for reproducibility available. Section \ref{sec:data_collection} describes in detail how both the real and synthetic data were collected and generated, and the complete dataset is publicly released through a permanent link (anonymous for now). The code used for training and evaluating all baseline detectors (including model weights) is provided alongside the \method and \textit{in-the-wild} datasets, ensuring that the reported experiments can be replicated. Appendix \ref{ap:details} gives further implementation details, including training procedures, hyperparameters, and compute resources. These resources should enable independent researchers to reproduce our results and extend the benchmarks under comparable settings.

\begin{ack}
This work was partially funded by the CIFAR AI Chairs Program. Additional financial support was provided by the Centre for the Study of Democratic Citizenship (CSDC), IVADO, and the Canada First Research Excellence Fund. We also gratefully acknowledge Mila for both financial assistance and access to computational resources. Finally, we thank Fuxiao Gao and Jie Zang for their help with data collection, as well as all participants who contributed to the human perception survey.
\end{ack}

\bibliographystyle{abbrvnat}
\bibliography{references}
\newpage
\appendix
\section{\textsc{OpenFake} composition and licensing}
\label{ap:stats}
\Cref{tab:dataset_stats} summarizes the dataset at the generator level, listing each base model and its LoRA or finetuned variants with release month, exact image counts, and license category. In total there are \textbf{963{,}342} synthetic images drawn from Stable Diffusion (1.5/2.1/XL/3.5), Flux (1.0 dev, 1.1 Pro, Schnell), Midjourney (6/7), DALL\textperiodcentered E~3, Imagen (3/4), GPT Image~1, Grok~2, Ideogram~3.0, HiDream, Chroma, and Recraft v3. The real corpus contains \textbf{3M} filtered LAION-400M images. Some proprietary and out-of-distribution generators appear with smaller totals because they were sourced from external collections rather than produced end-to-end.

Licensing and access are made explicit to support downstream compliance. We label sources as \emph{Community}, \emph{Non-commercial}, or \emph{Non-compete} and include these labels in the release metadata. All manifests are hosted on the HuggingFace Hub in streaming-friendly formats (Parquet and CSV) with per-item metadata such as model family, variant, release month, and prompt text.

\begin{table}[h]
    \centering
    \small
    \begin{tabular}{lrrr}
        \toprule
        \textbf{Source} & \textbf{Release (YYYY--MM)} & \textbf{\# Images} & \textbf{Licence} \\
        \midrule
        \textit{Real (LAION-400M, filtered)} & 2021--08 & $^*$3M & -- \\
        \midrule
        Stable Diffusion 1.5 & 2022--08 & 76{,}510 & Community \\
        \quad {\scriptsize Stable Diffusion 1.5 (base)} & {\scriptsize 2022--08} & {\scriptsize 20{,}000} & {\scriptsize Community} \\
        \quad {\scriptsize Dreamshaper (FT)} & {\scriptsize 2023--07} & {\scriptsize 36{,}510} & {\scriptsize Community} \\
        \quad {\scriptsize EpicDream (FT)} & {\scriptsize 2023--08} & {\scriptsize 20{,}000} & {\scriptsize Community} \\
        \midrule
        Stable Diffusion 2.1 & 2022--12 & 135{,}487 & Community \\
        \midrule
        Stable Diffusion XL & 2023--07 & 186{,}666 & Community \\
        \quad {\scriptsize Stable Diffusion XL (base)} & {\scriptsize 2023--07} & {\scriptsize 40{,}000} & {\scriptsize Community} \\
        \quad {\scriptsize Epic Realism (FT)} & {\scriptsize 2025--06} & {\scriptsize 59{,}770} & {\scriptsize Community} \\
        \quad {\scriptsize Touch of realism (LoRA)} & {\scriptsize 2025--06} & {\scriptsize 32{,}828} & {\scriptsize Community} \\
        \quad {\scriptsize RealVisXL-v5 (FT)} & {\scriptsize 2024--09} & {\scriptsize 29{,}300} & {\scriptsize Community} \\
        \quad {\scriptsize Juggernaut (FT)} & {\scriptsize 2025--05} & {\scriptsize 24{,}768} & {\scriptsize Community} \\
        \midrule
        Flux 1.0 dev & 2024--08 & 144{,}788 & Non-commercial \\
        \quad {\scriptsize Flux 1.0 dev (base)} & {\scriptsize 2024--08} & {\scriptsize 106{,}796} & {\scriptsize Non-commercial} \\
        \quad {\scriptsize Mystic (FT)} & {\scriptsize 2024--10} & {\scriptsize 15{,}608} & {\scriptsize Non-commercial} \\
        \quad {\scriptsize MVC5000 (LoRA)} & {\scriptsize 2025--07} & {\scriptsize 16{,}244} & {\scriptsize Non-commercial} \\
        \quad {\scriptsize Amateur Snapshot Photos (LoRA)} & {\scriptsize 2025--06} & {\scriptsize 4{,}140} & {\scriptsize Non-commercial} \\
        \quad {\scriptsize Realism (LoRA)} & {\scriptsize 2024--08} & {\scriptsize 2{,}000} & {\scriptsize Non-commercial} \\
        \midrule
        DALL\textperiodcentered E~3 & 2023--10 & 33{,}336 & Non-compete \\
        Midjourney 6 & 2023--12 & 50{,}000 & Non-compete \\
        Imagen 3.0 & 2024--08 & 4{,}032 & Non-compete \\
        Flux-1.0-Schnell & 2024--08 & 36{,}084 & Non-commercial \\
        Flux-1.1-Pro & 2024--10 & 29{,}923 & Non-commercial \\
        Recraft v3 & 2024--10 & 1{,}000 & Community \\
        Stable Diffusion 3.5 & 2024--10 & 139{,}114 & Non-compete \\
        Grok 2 & 2024--12 & 9{,}803 & Non-compete \\
        Ideogram 3.0 & 2025--03 & 28{,}495 & Non-compete \\
        Midjourney 7 & 2025--04 & 3586 & Non-compete \\
        GPT Image 1 & 2025--04 & 41{,}315 & Non-compete \\
        HiDream-I1 Full & 2025--04 & 27{,}904 & Community \\
        Imagen 4.0 & 2025--05 & 10{,}721 & Non-compete \\
        Chroma & 2025--08 & 4{,}532 & Community \\
        \midrule
        \textbf{Total synthetic} & -- & \textbf{963,342} & -- \\
        \bottomrule
    \end{tabular}
    \vspace{0.5em}
    \caption{\textsc{OpenFake} statistics. Image counts are exact. $^*$While we release the entire 3M real images and prompts, only a balanced subset is fully uploaded to the HuggingFace Hub to match the number of fake images. The remainder can be downloaded via URLs provided in CSV files on the Hub. LoRA variants (``LoRA'') and full finetunes (``FT'') are listed on separate, smaller rows directly below their base models.}
    \label{tab:dataset_stats}
\end{table}

\section{More results}
\subsection{Cross-benchmark generalization of SwinV2}
\label{ap:cross_bench}

\paragraph{Summary.}
\Cref{tab:swin_cross_bench} compares SwinV2 detectors trained on three datasets and evaluated across two external test suites. The \textsc{OpenFake}-trained model attains the best out-of-domain balance between TPR and TNR on both \textsc{GenImage} and \textsc{Semi-Truths}, translating into stronger Accuracy and F1. In-domain results (italicized) saturate, as expected, but are less informative about generalization.

\begin{table}[h]
\centering
\small
\setlength{\tabcolsep}{6pt}
\begin{tabular}{llccc}
\toprule
\textbf{Test set} & \textbf{Metric} &
\makecell{\textbf{Train}\\\textbf{OpenFake}} &
\makecell{\textbf{Train}\\\textbf{GenImage}} &
\makecell{\textbf{Train}\\\textbf{Semi-Truths}} \\
\midrule
\multirow{5}{*}{\textsc{GenImage}}
 & TPR          & 0.771 & \textit{1.000} & 0.965 \\
 & TNR          & 0.928 & \textit{1.000} & 0.261 \\
 & Accuracy     & 0.849 & \textit{1.000} & 0.613 \\
 & F1 Score     & 0.836 & \textit{1.000} & 0.714 \\
\midrule
\multirow{5}{*}{\textsc{Semi-Truths}}
 & TPR          & 0.909 & 0.830 & \textit{1.000} \\
 & TNR          & 0.962 & 1.000 & \textit{1.000} \\
 & Accuracy     & 0.920 & 0.865 & \textit{1.000} \\
 & F1 Score     & 0.947 & 0.907 & \textit{1.000} \\
\bottomrule
\end{tabular}
\vspace{0.5em}
\caption{Cross-benchmark generalization of SwinV2 detectors. Italicised numbers indicate \emph{in-domain} evaluations, where the model is tested on the same dataset it was trained on.  TPR = true-positive rate (synthetic images), TNR = true-negative rate (real images). All values are shown to three decimal places.}
\label{tab:swin_cross_bench}
\end{table}

\begin{figure}[h]
    \centering
    \vspace{-1em}
    \includegraphics[width=0.95\linewidth]{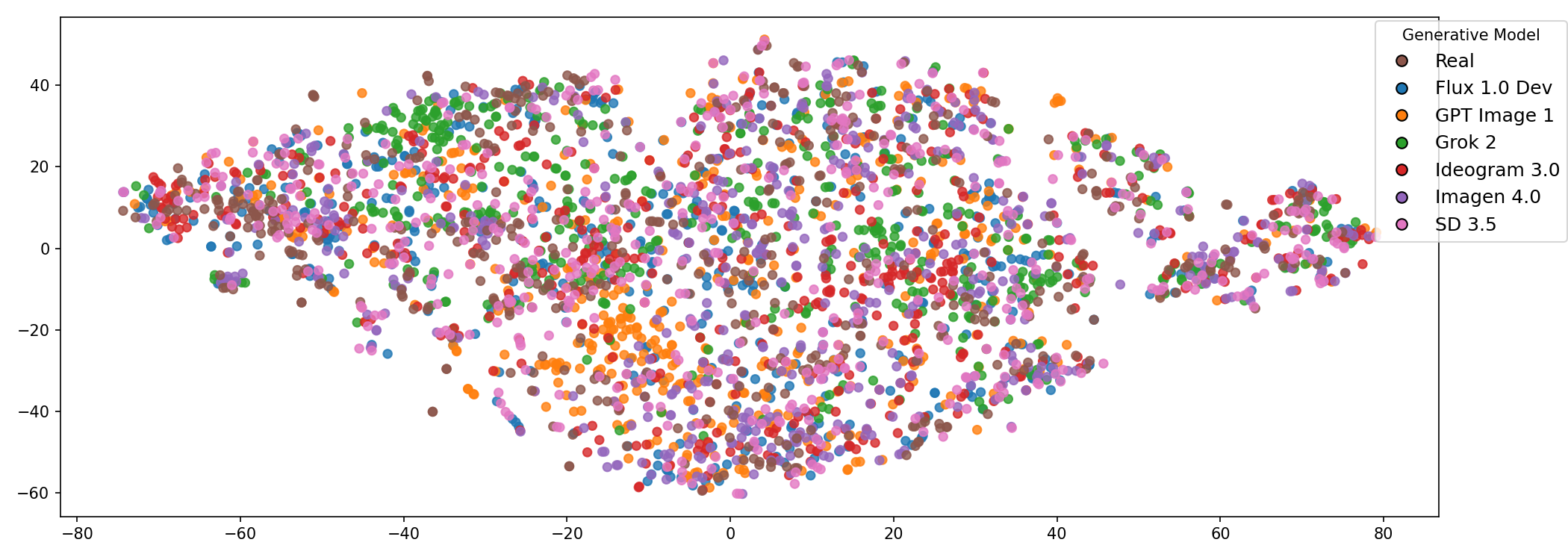}
    \caption{t-SNE visualization of CLIP vision embeddings for 3,500 test images, including both real and synthetic images from a few generative models. Each point corresponds to an individual image, and colours indicate the generative model (or “real” for authentic images).}
    \label{fig:tsne}
\end{figure}

\paragraph{Why \textsc{OpenFake} transfers better.}
We attribute the gains primarily to coverage and recency. \textsc{OpenFake} aggregates diverse, up-to-date generators and visual conditions (including compressions and photorealistic prompts), reducing shortcut reliance. By contrast, models trained on older or narrower distributions tend to overfit curation artifacts, which explains the high TPR but poor TNR observed when \textsc{Semi-Truths}-trained detectors face newer datasets: many real images are misclassified as synthetic.

\paragraph{Error patterns and operational trade-offs.}
The cross-benchmark gaps underscore the importance of calibrating for deployment goals. A detector with inflated TPR but depressed TNR can look strong on superficially balanced metrics yet cause unacceptable false-positive rates in real pipelines. Threshold selection, confidence calibration, and cost-sensitive training are therefore critical when transferring across domains.

\subsection{Robustness to compression artifacts}
\label{ap:artifacts}
Real images (sourced from LAION-400M in our dataset) are typically compressed and carry authentic JPEG artifacts and blur. In contrast, synthetic images are high-resolution and minimally compressed. This mismatch creates an obvious signal: detectors may rely on compression differences instead of true semantic features, thus failing on compressed fakes. To assess this vulnerability, we implemented a data augmentation pipeline to degrade synthetic images during training, mimicking the distribution of real images, as done in previous work \citep{corvi2023detection, Wang_2021_ICCV, cheng2024diffusion}. This includes random resizing, Gaussian blur, JPEG compression, and Gaussian noise. We then evaluated the SwinV2 model trained with these augmentations on a fully compressed test set and got an overall F1 score of 0.992. This demonstrates that the model remains highly accurate even when the compression signal is neutralized.
\section{Filtering and Captioning of LAION Images}
\label{ap:filtering}

To curate a relevant subset of real images from LAION-400M, we implemented a two-stage filtering and captioning pipeline using the vision-language model Qwen2.5-VL. This approach allowed us to filter politically salient and emotionally impactful content while preserving real-world visual characteristics (e.g., compression artifacts) crucial for training robust deepfake detectors.

\paragraph{Filtering prompt.}
The first step used a vision-language reasoning prompt to assess whether each image depicted (i) real human faces, and/or (ii) politically or emotionally significant events. Many original LAION captions are noisy or incomplete, so the model was asked to jointly analyze both image and caption. The prompt was:

\begin{quote}
\small
\ttfamily
Analyze the provided image and its caption: “\textit{\{caption\}}”.
Provide detailed reasoning on the following two points:
	
    1.	Does the image contain any real human face(s)? Exclude animations, cartoons, figurines, statues, drawings, paintings, or video games.
	
    2.	Does the image contain content related to political events, catastrophes, news events, or anything likely to have high emotional impact or polarization? Exclude animations, cartoons, drawings, paintings, or video games.

Conclude clearly with either “Humans: yes” or “Humans: no”, and “Catastrophes: yes” or “Catastrophes: no”.
\end{quote}

Only images with at least one “yes” label (human or catastrophe) were retained. This strategy allowed us to target both portrait-based and event-based misinformation vectors while filtering out non-photographic and low-impact content.

\paragraph{Captioning prompt.}
For the selected images, we generated improved prompts to guide synthetic image generation. These prompts describe the image in a style suitable for text-to-image models, incorporating visual format and subject matter. The Qwen2.5-VL prompt used was:

\begin{quote}
\small
\ttfamily
Given the image and its caption: “\textit{\{caption\}}”, generate a concise prompt in a single sentence that describes the image and its format (e.g., photograph, poster, screenshot), including any people present.
Do not mention the caption directly.
\end{quote}

These refined prompts were used for synthetic image generation and are also included in the public release to support downstream research and reproducibility.

\section{Ethics, Privacy \& Limitations}
\label{ap:limitations}

While our dataset aims to support robust deepfake detection, it inherits limitations from its sources. The real image corpus, derived from the LAION crawl (2014–2021), skews toward Western-centric and pre-pandemic imagery. Proprietary generative models also reflect aesthetic and cultural biases from their training data. These imbalances may affect the generalizability of detection models across diverse global contexts. We document these issues in the HuggingFace Data Card and encourage contributions from underrepresented regions via our Arena pipeline. 

The paper includes details of both the human perception study and the Arena crowdsourcing platform. No compensation was offered, as participation was voluntary, and both systems were designed to ensure anonymity and avoid the collection of personal data.


Prompt extraction may introduce semantic noise, and the quality of adversarial data depends on user participation. Our dataset focuses on visual realism, but does not yet capture multimodal or context-based misinformation. Fairness across demographic groups and long-term robustness remain open challenges. We encourage downstream audits and broader evaluation to support responsible deployment.

\section{Training Details and Compute Resources}
\label{ap:details}

\subsection{Compute Resources and Cost}
\label{ap:cost}
All experiments were conducted on an internal compute cluster or local workstations with moderate storage and GPU availability. Below, we detail the computational resources and costs associated with dataset filtering, image generation, baseline evaluation, and dataset hosting.

\paragraph{Filtering and analysis.} The LAION filtering pipeline ran continuously for two weeks on 4 NVIDIA L40S GPUs (48 GB VRAM each). An additional 2 days of compute on the same setup was used for prompt selection and vision–language model evaluation, comparing multiple candidate models and prompt formats.

\paragraph{Synthetic image generation.} Images from Stable Diffusion v2.1 and Flux.1.0-dev were generated on 4 L40S GPUs over a span of 4 days per model. Other models generated images for 1 day. Each GPU was fully utilized to maximize throughput.

\paragraph{Model training and evaluation.} Training the SwinV2 baseline classifier on the \textsc{OpenFake} dataset required approximately 12 hours on a single NVIDIA L40S GPU. Inference for evaluation purposes was negligible in comparison.

\paragraph{Baseline inference.} For baseline evaluation:
\begin{itemize}
    \item InternVL inference over the full test set was performed over 10 hours on a single RTX8000 GPU (48 GB VRAM).
    \item CLIP and the Corvi2023 baselines were evaluated in approximately 6 hours on the same RTX8000 GPU.
\end{itemize}

\paragraph{Proprietary model generation.} Images generated via proprietary APIs incurred a per-image cost of approximately \$0.04 (USD), varying slightly by model and resolution. No GPU compute was required on our end; generation was offloaded entirely to the remote API services.

\paragraph{Storage and hosting.} Dataset preprocessing, metadata formatting, and uploads to Hugging Face required only CPU cores but substantial storage capacity. The working set size during dataset preparation exceeded 1TB.

\textbf{Total estimated GPU compute:} $\sim$4 GPU-months across L40S and RTX8000 class cards. All compute was performed on institutional resources without incurring cloud costs.

\subsection{SwinV2 Fine-Tuning Hyperparameters}
\label{app:hyperparams}

For our main benchmark detector, we finetune \textit{microsoft/swinv2-small-patch4-window16-256} on the \textsc{OpenFake} dataset using the HuggingFace \texttt{Trainer} API. All experiments were conducted on a single L40S GPUs.

\paragraph{Model architecture.} We use the SwinV2-Small transformer backbone with the classifier head modified to predict two classes: real vs. fake. The model is initialized from ImageNet-1k weights and fine-tuned end-to-end.

\paragraph{Input resolution.} Images are resized to $256 \times 256$ using the default SwinV2 image processor.

\paragraph{Training configuration.}
\begin{itemize}
\item \textbf{Optimizer:} AdamW
\item \textbf{Learning rate:} 5e-5
\item \textbf{Batch size:} 32
\item \textbf{Epochs:} 5
\item \textbf{Learning rate scheduler:} Linear with warmup
\end{itemize}

\paragraph{Data augmentation.}
During training we use two augmentation streams. A general geometric/photometric stream is applied to \emph{both} real and synthetic images, including random resized crops, color jitter, small rotations, occasional horizontal flips, and mild Gaussian blur. To neutralize compression shortcuts, a light \emph{degradation} stream is applied to \emph{synthetic} images only, including resolution downscaling, blur adjustment, low-level Gaussian noise, and JPEG compression with randomized quality. Transforms are sampled stochastically, and the synthetic-only degradations are calibrated to match statistics of LAION-derived real images. For compressed test-set evaluation, synthetic images are post-processed with the same degradation function to simulate internet-style artifacts; we report accuracy, precision, recall, F1, and ROC AUC.

\subsection{Generation Parameters for Open-Source Models}
\label{app:gen_params}

We document here the generation settings used to produce synthetic images from open-source models within the \textsc{OpenFake} dataset. This ensures reproducibility and clarity on the diversity of generated outputs.

We used \texttt{stabilityai/stable-diffusion-3.5-large} to generate synthetic images and \texttt{black-forest-labs/Flux.1.0-dev} using the same bank of prompts. Both models were run in \texttt{bfloat16} precision using their official pipelines—\texttt{StableDiffusion3Pipeline} and \texttt{FluxPipeline}, respectively—and deployed across multiple GPUs with prompt sharding and batched inference for scalability. We used the official HugginFace weights for the other models via the \texttt{Diffusers} Python library.

For all models, the following generation settings were generally applied (there could be slight modifications based on the recommended parameters for each model):

\begin{itemize}
\item \textbf{Resolution}: Randomly sampled from a predefined set of social-media-style sizes: \texttt{[(1024, 1024), (1024, 512), (512, 1024), (1024, 768), (768, 1024), (1152, 768), (768, 1152)]}
\item \textbf{Guidance scale}: Uniformly sampled between 1.5 and 7
\item \textbf{Inference steps}: [10, 40]
\item \textbf{Scheduler}: Default
\end{itemize}

These configurations were chosen to maximize diversity and photorealism, while reflecting the resolution and stylistic variability typical of online content.

\section{\textsc{OpenFake Arena}}
\label{ap:arena}
\begin{figure}[ht]
\centering
\includegraphics[width=0.99\linewidth]{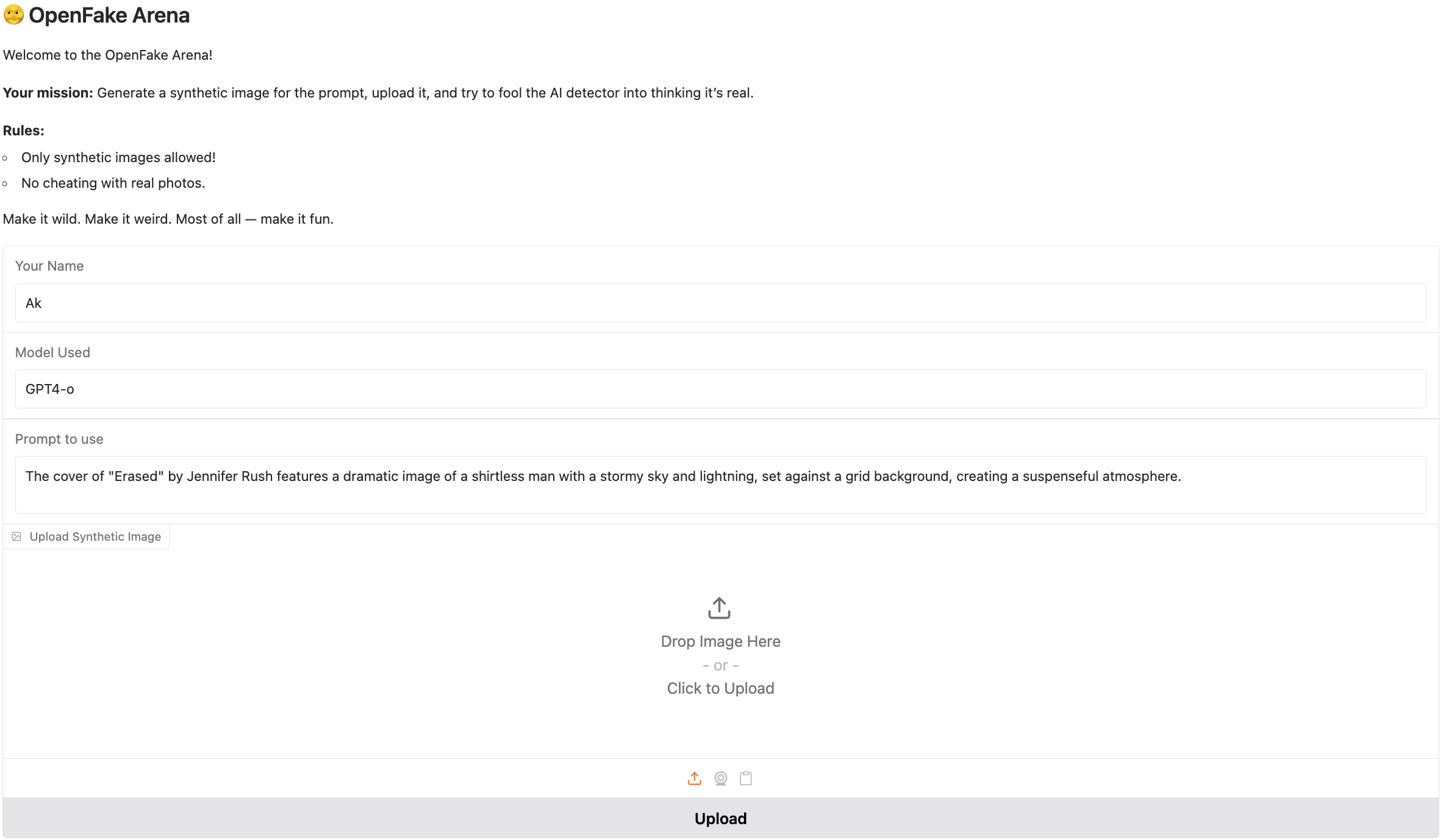}
\caption{\textsc{OpenFake Arena} interface. Users are presented with a prompt and asked to generate an image that can fool the detector.}
\label{fig:arena1}
\end{figure}

\begin{figure}[ht]
\centering
\includegraphics[width=0.99\linewidth]{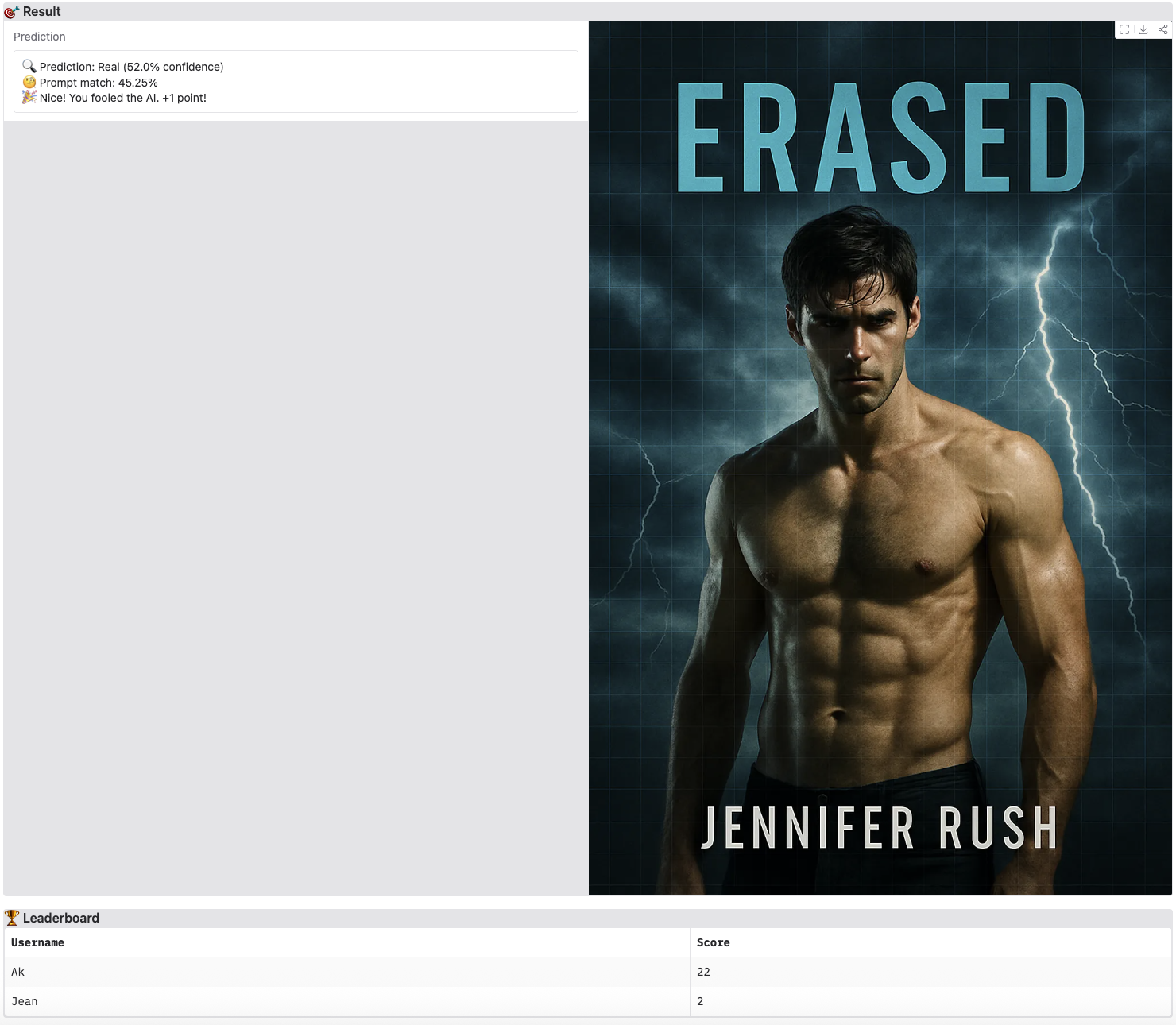}
\caption{Example of a successful submission. The image aligns with the prompt "The cover of "Erased" by Jennifer Rush features a dramatic image of a shirtless man with a stormy sky and lightning, set against a grid background, creating a suspenseful atmosphere". It is incorrectly classified as real by the detector.}
\label{fig:arena3}
\end{figure}

\begin{figure}[ht]
\centering
\includegraphics[width=0.99\linewidth]{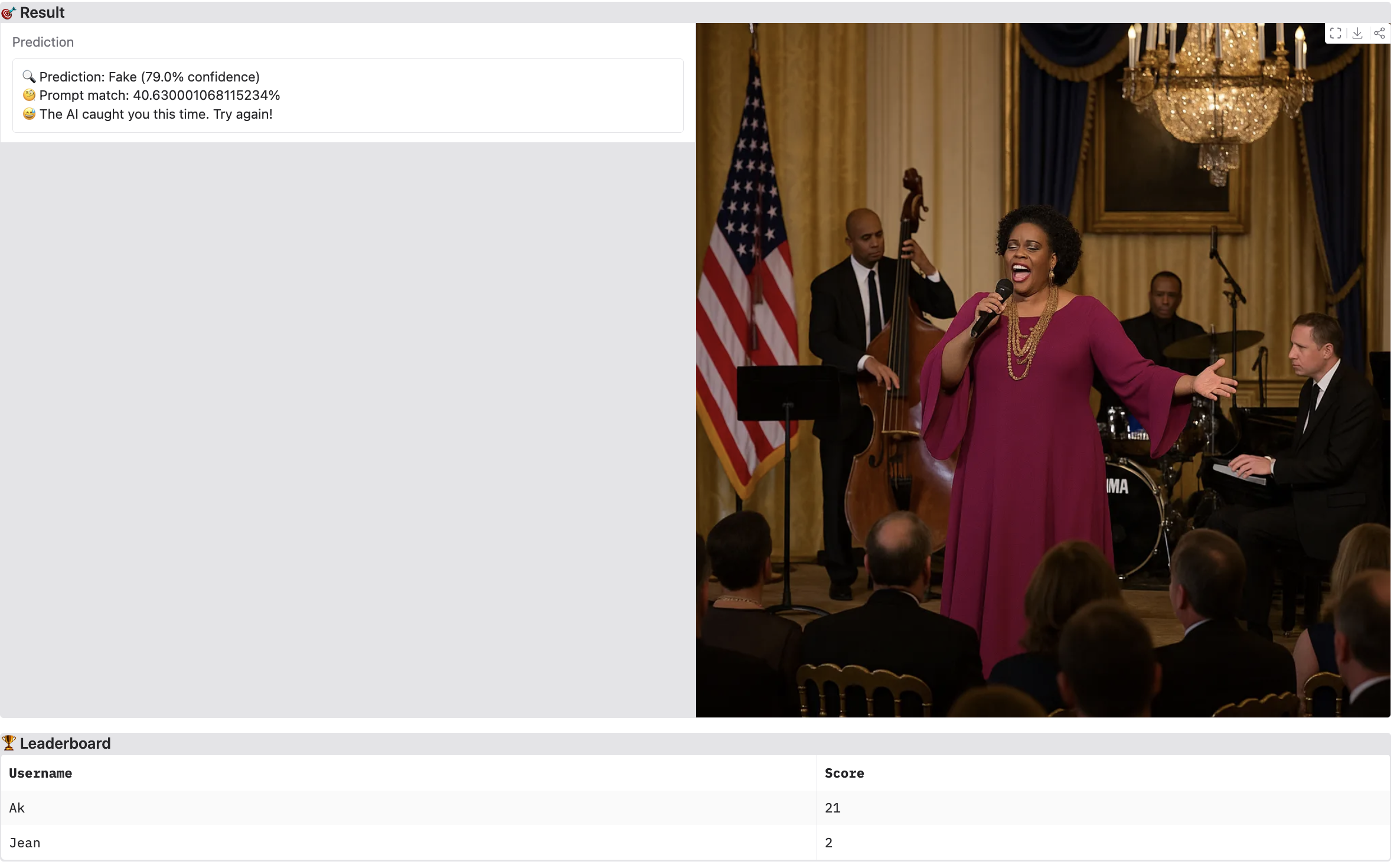}
\caption{Example of an unsuccessful submission. The image fails to fool the detector and is correctly classified as synthetic. The prompt used was "A photograph captures Dianne Reeves performing on stage in the East Room of the White House during the National Governors Association Dinner on February 26, 2012, with an audience seated in the foreground."}
\label{fig:arena2}
\end{figure}

We host the Arena as a Gradio app on Hugging Face Spaces, leveraging their compute resources. A pretrained CLIP model acts as a prompt-matching gate to ensure image relevance, and successful submissions that fool the detector are stored in a connected Hugging Face dataset. The detector is a SwinV2 model trained on the \textsc{OpenFake} dataset and periodically updated to reflect new data. We also log metadata such as the generative model used and the user ID to support leaderboard tracking. Prompts are designed to be specific and difficult to spoof, and additional safeguards are in place to prevent misuse. Upon acceptance, we plan to promote the Arena through social media and at the conference to encourage broader participation. Figures \ref{fig:arena1}, \ref{fig:arena2}, and \ref{fig:arena3} show the Arena interface and leaderboard, along with examples of successful and failed submissions.

\FloatBarrier
\section{Synthetic image examples from \textsc{OpenFake}}

\begin{figure}[ht!]
  \centering
  \foreach[count=\n] \i in {0,1,2,3,4,5,6,7,8,9,10,11,12,14,15,17}{%
      \includegraphics[height=2.3cm]{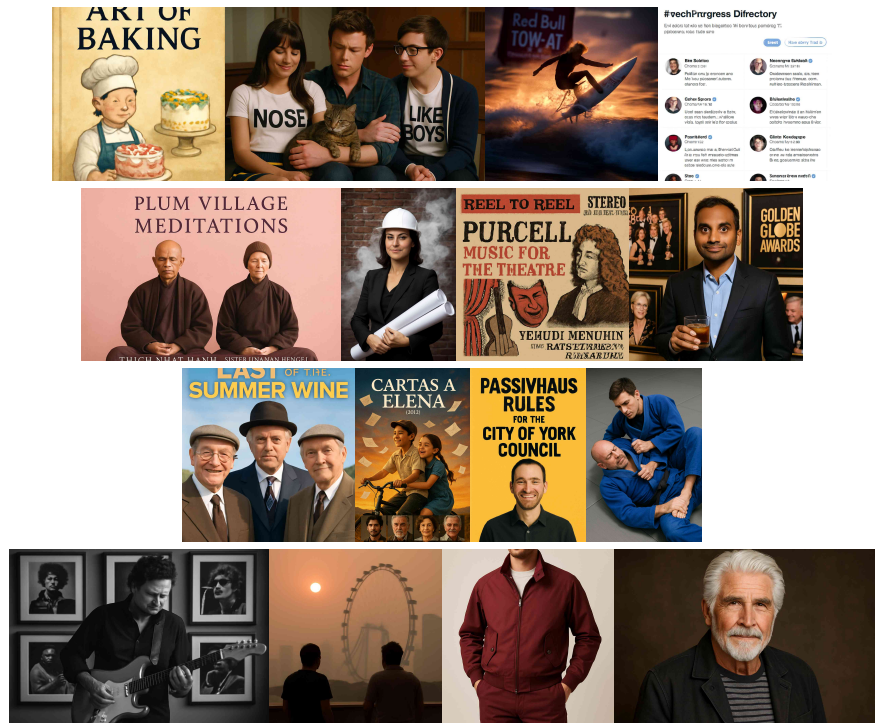}%
      \ifnum\n=4 \\[2pt] \fi%
      \ifnum\n=8 \\[2pt] \fi%
      \ifnum\n=12 \\[2pt] \fi%
  }
  \caption{Sample images from \textsc{OpenFake} generated by \textbf{GPT Image 1}.}
  \label{fig:gpt_examples}
\end{figure}

\begin{figure}[h!]
  \centering
  \foreach[count=\n] \i in {1,2,3,4,5,6,7,8,9,10,11,19,13,14,15,18}{%
      \includegraphics[height=2.3cm]{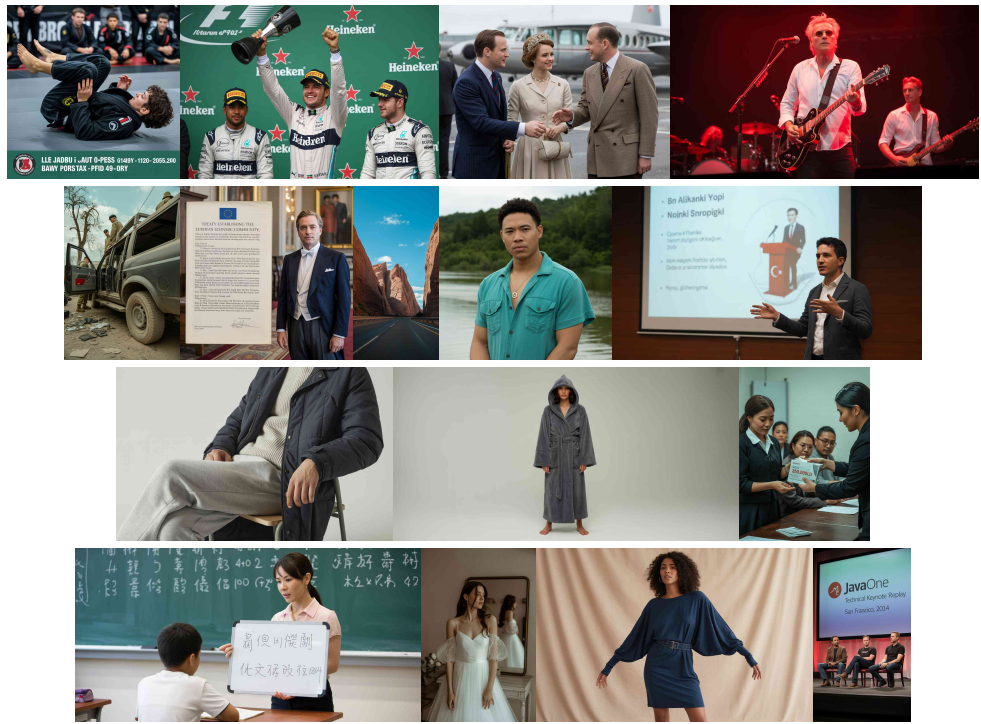}%
      \ifnum\n=4 \\[2pt] \fi%
      \ifnum\n=9 \\[2pt] \fi%
      \ifnum\n=12 \\[2pt] \fi%
  }
  \caption{Sample images from \textsc{OpenFake} generated by \textbf{Ideogram 3.0}.}
  \label{fig:ideogram_examples}
\end{figure}

\begin{figure}[h]
  \centering
  \foreach[count=\n] \i in {2,4,5,6,7,8,9,10,11,12,13,14,15,16,17,18}{%
      \includegraphics[height=2.3cm]{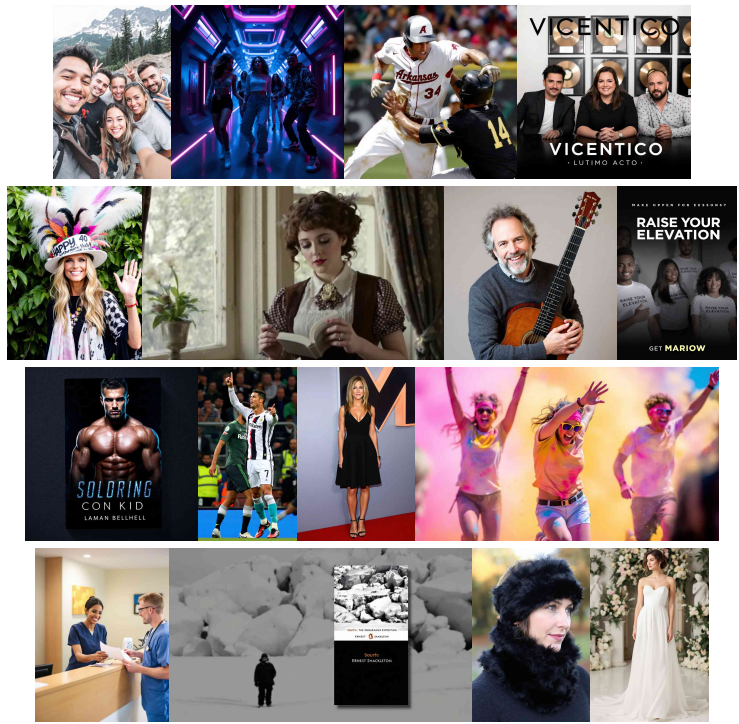}%
      \ifnum\n=4 \\[2pt] \fi%
      \ifnum\n=8 \\[2pt] \fi%
      \ifnum\n=12 \\[2pt] \fi%
  }
  \caption{Sample images from \textsc{OpenFake} generated by \textbf{Flux-1.1‑Pro}.}
  \label{fig:fluxpro_examples}
\end{figure}

\begin{figure}[h!]
  \centering
  \foreach[count=\n] \i in {0,1,2,3,4,5,6,7,8,9,10,11,12,13,14,15}{%
      \includegraphics[height=2.3cm]{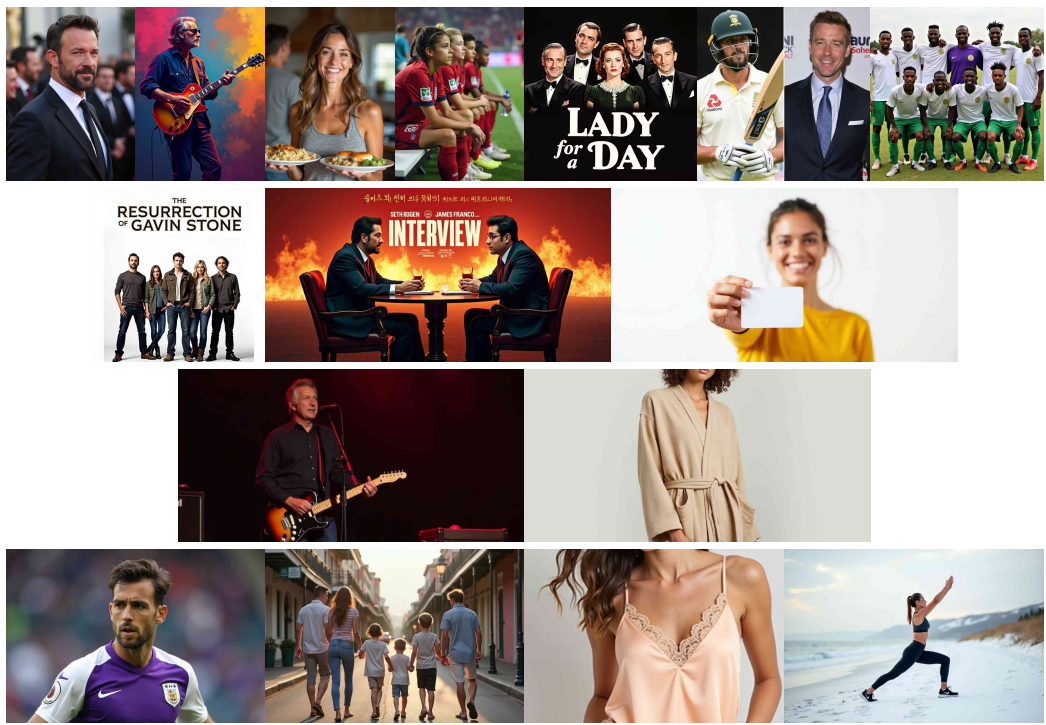}%
      \ifnum\n=7 \\[2pt] \fi%
      \ifnum\n=10 \\[2pt] \fi%
      \ifnum\n=12 \\[2pt] \fi%
  }
  \caption{Sample images from \textsc{OpenFake} generated by \textbf{Flux.1‑Dev}.}
  \label{fig:fluxdev_examples}
\end{figure}

\begin{figure}[h!]
  \centering
  \foreach[count=\n] \i in {0,1,2,3,4,5,6,7,8,9,10,11,12,13,14,15}{%
      \includegraphics[height=2.3cm]{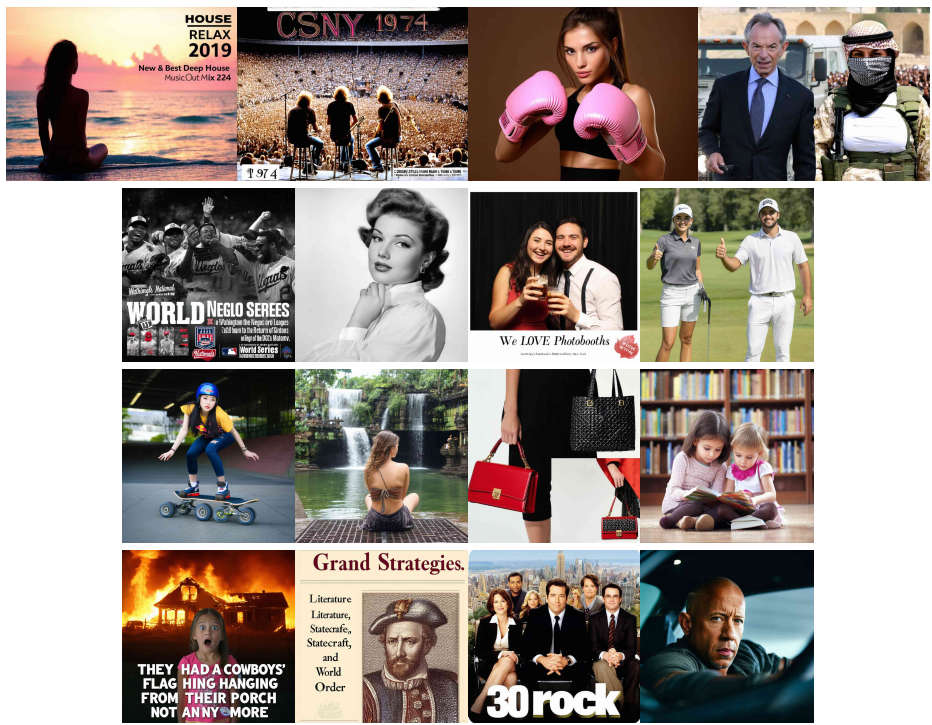}%
      \ifnum\n=4 \\[2pt] \fi%
      \ifnum\n=8 \\[2pt] \fi%
      \ifnum\n=12 \\[2pt] \fi%
  }
  \caption{Sample images from \textsc{OpenFake} generated by \textbf{Stable Diffusion 3.5}.}
  \label{fig:sd3_examples}
\end{figure}

\FloatBarrier

\end{document}